\documentclass[11pt,a4paper]{article}
\usepackage[hyperref]{acl2020}
\usepackage{times}
\usepackage{latexsym}

\usepackage{enumitem}
\usepackage{multirow}
\usepackage{booktabs}
\usepackage{graphics}
\usepackage{graphicx}
\usepackage{subfigure}
\usepackage{tabularx}

\definecolor{berry}{HTML}{BC3754}

\usepackage{microtype}

\aclfinalcopy 
\def\aclpaperid{362} 

\title{Moving Down the Long Tail of Word Sense Disambiguation\\ with Gloss Informed Bi-encoders}

\author{Terra Blevins and Luke Zettlemoyer \\
        Paul G. Allen School of Computer Science \& Engineering, University of Washington \\
        Facebook AI Research, Seattle \\
        {\tt \{blvns, lsz\}@cs.washington.edu}}

\date{}

\begin{document}
\maketitle
\begin{abstract}
A major obstacle in Word Sense Disambiguation (WSD) is that word senses are not uniformly distributed, causing existing models to generally perform poorly on senses that are either rare or unseen during training. We propose a bi-encoder model that independently embeds (1) the target word with its surrounding context and (2) the dictionary definition, or gloss, of each sense.
The encoders are jointly optimized in the same representation space, so that sense disambiguation can be performed by finding the nearest sense embedding for each target word embedding.
Our system outperforms previous state-of-the-art models on English all-words WSD; these gains predominantly come from improved performance on rare senses, leading to a 31.1\% error reduction on less frequent senses over prior work. This demonstrates that rare senses can be more effectively disambiguated by modeling their definitions.
\end{abstract}

\section{Introduction}
One of the major challenges of Word Sense Disambiguation (WSD) is overcoming the data sparsity that stems from the Zipfian distribution of senses in natural language \cite{kilgarriff2004dominant}. For example, in SemCor (the largest manually annotated dataset for WSD) 90\% of mentions of the word \textit{plant} correspond to the top two senses of the word, and only half of the ten senses of \textit{plant} occur in the dataset at all \cite{miller1993semantic}. Due to this data imbalance, many WSD systems show a strong bias towards predicting the most frequent sense (MFS) of a word regardless of the surrounding context~\cite{postma2016more}.

A successful WSD system should be able to overcome this bias and correctly disambiguate cases where a word takes a less frequent sense (LFS), without sacrificing performance on MFS examples. Previous work has found that incorporating lexical information such as sense definitions, or glosses, into WSD systems improves performance \cite{luo2018leveraging, luo2018incorporating}.\footnote{For example, in the sentence ``She \textit{planted} the tree," the gloss, or meaning, for the sense of \textit{plant} is ``put or set [something] firmly into the ground." \cite{miller1995wordnet}} 
Glosses have also been found to improve LFS performance; however, absolute performance on rare senses is still low, with models showing a 62.3 F1 performance drop between the MFS examples and the LFS ones \cite{kumar2019zero}.

In this paper, we show that this gap can be significantly reduced by jointly fine-tuning multiple pretrained encoders on WSD. We present a bi-encoder model built on top of BERT \cite{devlin2019bert} that is designed to improve performance on rare and zero-shot senses. Similar to prior work, our system represents the target words and senses in the same embedding space by using a \textit{context encoder} to represent the target word and surrounding context, and a \textit{gloss encoder} to represent the sense definitions. However, our two encoders are jointly learned from the WSD objective alone and trained in an end-to-end fashion.

This approach allows our model to outperform prior work on the English all-words WSD task introduced in \citet{raganato2017word}. Analysis of our model shows that these gains come almost entirely from better performance on the less frequent senses, with an 15.6 absolute improvement in F1 performance over the closest performing system; our model also improves on prior work in the zero-shot setting, where we evaluate performance on words and senses not seen during training.

Finally, we train our model in a few-shot setting in order to investigate how well the bi-encoder system learns on a limited set of training examples per sense. The bi-encoder architecture is able to generalize better from the limited number of examples than a strong pretrained baseline. This results demonstrates the data efficiency of our system and indicates why it captures LFS well, as less common senses naturally only have a few training examples in the data.

In summary, the overall contributions of this work are as follows:
\begin{itemize}[topsep=1pt,itemsep=-1pt,partopsep=4pt, parsep=4pt]
    \item We present a jointly optimized bi-encoder model (BEM) for WSD that improves performance on all-words English WSD.
    \item We show that our model's improvements come from better performance on LFS and zero-shot examples, without sacrificing accuracy on the most common senses.
    \item We examine why our model performs well on LFS with a number of experiments, including an evaluation of the BEM in a few-shot learning setting demonstrating that the bi-encoder generalizes well from limited data.
\end{itemize}
The source code and trained models for our WSD bi-encoders can be found at \url{https://github.com/facebookresearch/wsd-biencoders}.

\begin{figure*}[t!]
        \includegraphics[width=\textwidth]{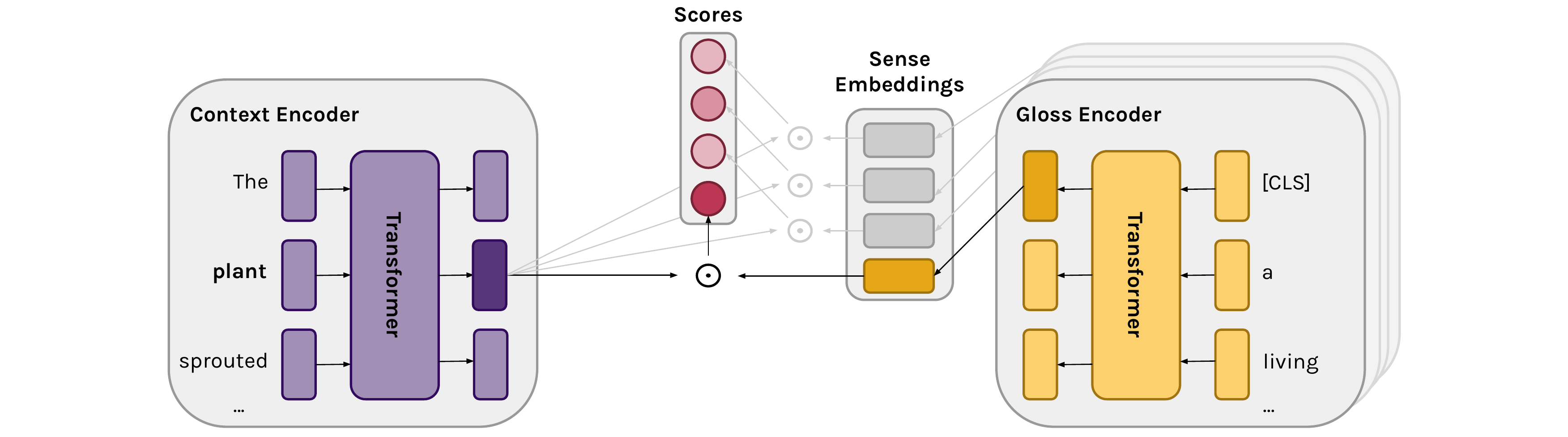}
        \caption{Architecture of our bi-encoder model for WSD. The context sentence and sense gloss text are input into the context and gloss encoders, respectively; each encoder is initialized with BERT. We take the $i^{th}$ output of the context encoder as the representation for the target word $w_i$; the first output of the gloss encoder, which corresponds to the BERT-specific start token [CLS], is used as a representation for each candidate sense $s$. $w_i$ is compared to $s$ with a dot product, and the sen
        se with the highest similarity to $w_i$ is assigned as the predicted label.}
        \label{architecture-fig}
\end{figure*}

\section{Background and Related Work}
Word Sense Disambiguation (WSD) is the task of predicting the particular sense, or meaning, of a word when it occurs in a specific context \cite{navigli2009word}. Understanding what a word means in context is critical to many NLP tasks, and WSD has been shown to help downstream tasks such as machine translation (MT) \cite{vickrey2005word, neale2016word,  rios-gonzales2017improving} and information extraction (IE) \cite{ciaramita2006broad, bovi2015knowledge}.

The formulation of WSD that we address is all-words WSD, where the model disambiguates every ambiguous word in the data (e.g., \citet{palmer2001english, moro2015semeval}). Many WSD systems approached this task with manually engineered features that were used to learn an independent classifier, or \textit{word expert}, for each ambiguous lemma \cite{zhong2010makes, shen2013coarse}. Later work also integrated word embeddings into this independent classifier approach \cite{rothe2015autoextend, iacobacci2016embeddings}. 

Neural models for WSD built on this approach by training encoders for better feature extraction; they then either still learned independent classifiers on top of the encoded features \cite{kaageback2016word}, or labeled each word using a shared output space \cite{raganato2017neural}. Other neural approaches used semi-supervised learning to augment the learned representations with additional data \cite{melamud2016context2vec, yuan2016semi}.

\subsection{Lexical Resources for WSD}
Definitions of senses, or glosses, have been shown to be a valuable resource for improving WSD. \citet{lesk1986automatic} used the overlap between the definitions of senses and the context of the target word to predict the target sense. This approach was later extended to incorporate WordNet graph structure \cite{banerjee2003extended} and to incorporate word embeddings \cite{basile2014enhanced}. More recently, \citet{luo2018leveraging, luo2018incorporating} added sense glosses as additional inputs into their neural WSD system, significantly improving overall performance.

Most similar to our work, \citet{kumar2019zero} represented senses as continuous representations learned from encoded glosses. However, they took a pipelined approach and supervised the gloss encoder with knowledge graph embeddings; they then froze the sense representations to use them as static supervision for training the WSD system. This approach requires an additional form of supervision (for which they used knowledge graph embeddings), making it more difficult to generalize to new data without that source of supervision. In comparison, our model is trained in an end-to-end manner and learns to embed gloss text without additional supervision.

Other work has shown that neural models capture useful semantic information about words from their definitions, and has used them to encode lexical representations \cite{bahdanau2017learning, bosc2018auto}. While they focused on representing words, rather than specific senses, their modeling approaches could be extended to sense representations.

\subsection{Pretrained NLP Models for WSD}
Pretrained models have been shown to capture a surprising amount of word sense information from their pretraining objectives alone \cite{peters2018deep, stanovsky2018spot, coenen2019visualizing}, allowing the frozen pretrained representations to compete with previous state-of-the-art WSD systems \cite{hadiwinoto2019improved}. Building on these findings, \citet{vial2019sense} incorporates pretrained BERT representations as inputs into their WSD system, and  \citet{loureiro2019language} uses BERT's contextualized outputs to create sense embeddings for each sense in WordNet.

Another approach to using pretrained models for WSD is to formulate the task as a sentence-pair classification problem, in which (context sentence, gloss) pairs are concatenated and cross-encoded with the pretrained model. This reduces the WSD task to a binary classification problem where the model is trained to predict whether the gloss matches the sense of the target word in the context sentence \cite{huang2019glossbert}. Given that transformer compute scales polynomially in the input length, our approach of independently encoding the contexts and sense glosses is more computationally efficient, and we also show that it performs better on the all-words WSD task (Section \ref{overall-results-section}).

\section{Methodology}
In this section, we present an approach for WSD that is designed to more accurately model less frequent senses by better leveraging the glosses that define them. The overall model architecture is shown in Figure \ref{architecture-fig}. Our bi-encoder model (BEM) consists of two independent encoders: (1) a \textit{context encoder}, which represents the target word (and its surrounding context) and (2) a \textit{gloss encoder}, that embeds the definition text for each word sense. These encoders are trained to embed each token near the representation of its correct word sense. 
Each encoder is a deep transformer network initialized with BERT, in order to leverage the word sense information it captures from pretraining \cite{coenen2019visualizing, hadiwinoto2019improved}. To describe our approach, we formally define the task of WSD (Section \ref{wsd-definition-section}), and then present the BEM system in detail (Section \ref{bem-section}).

\subsection{Word Sense Disambiguation}
\label{wsd-definition-section}
Word Sense Disambiguation (WSD) is the task of assigning a sense to a target word, given its context. More formally, given a word $w$ and context $c$, a WSD system is a function $f$ such that $f(w, c) = s$ subject to $s \in S_{w}$, where $S_{w}$ is all possible candidate senses of $w$.

We focus on the task of all-words WSD, in which every ambiguous word in a given context is disambiguated.\footnote{In practice, this means every \textit{content} word -- noun, verb, adjective, and adverb -- in the context is disambiguated by the WSD system.} In this setting, a WSD model is given as input $\mathbf{c} = c_0, c_1, ..., c_n$ and outputs a sequence of sense predictions $\mathbf{s} = s_{c_0}^{i}, s_{c_1}^j, ..., s_{c_n}^m$, where the model predicts the $i^{th}$, $j^{th}$, and $m^{th}$ senses from the candidate sense sets for $c_0$, $c_1$, and $c_n$, respectively. For our approach, we assume for each sense $s$ that we also have a gloss $g_s = g_0, g_1,...,g_n$ that defines $s$.

\subsection{Bi-encoder for WSD}
\label{bem-section}
Our bi-encoder architecture independently encodes target words (with their contexts) and sense glosses \cite{bromley1994signature, humeau2019real}. Each of these models are initialized with BERT-base: therefore, the inputs to each encoder are padded with BERT-specific start and end symbols: input $\mathbf{z} = z_0, z_1, ..., z_n$ is modified to $\mathbf{z} = $[CLS], $z_0, z_1, ..., z_n,$ [SEP].

The \textbf{context encoder}, which we define as $T_c$, takes as input a context sentence $\mathbf{c}$ containing a set of target words $\mathbf{w}$ to be disambiguated, s.t. $\mathbf{c} = c_0, c_1,...,w_i,...,c_n$, where $w_i$ is the $i^{th}$ target word in the context sentence. The encoder then produces a sequence of representations $\mathbf{r}$, where 
$$r_{w_i} = T_c(\mathbf{c})[i]$$
or the $i^{th}$ representation output by $T_c$. For words that are tokenized into multiple subword pieces by the BERT tokenizer, we represent the word by the average representation of its subword pieces. For example, let the $j^{th}$ through $k^{th}$ tokens correspond to the subpieces of the $i^{th}$ word, we have
$$r_{w_i} = \frac{1}{k-j} \sum_{l=j}^{k} (T_c(\mathbf{c})[l])$$

The \textbf{gloss encoder}, defined as $T_g$, takes in a gloss $\mathbf{g_s} = g_0, g_1, ..., g_m$ that defines the sense $s$ as input. The gloss encoder represents $s$ as 
$$r_s = T_g(\mathbf{g_s})[0]$$ 
where we take the first representation output by the gloss encoder (corresponding to the input [CLS] token) as a global representation for $s$. 

We then \textbf{score} each candidate sense $s \in S_w$ for a target word $w$ by taking the dot product of $r_w$ against every $r_s$ for $s \in S_w$: 
$$\phi(w, s_i) = r_w \cdot r_{s_i}$$ 
for $i = 0,...,|S_w|$. During evaluation, we predict the sense $\hat{s}$ of the target word $w$ to be the sense $s_i \in S_w$ whose representation $r_{s_i}$ has the highest dot product score with $r_w$.

We use a \textbf{cross-entropy loss} on the scores for the candidate senses of the target word $w$ to train our bi-encoder model; the loss function of our system given a (word, sense) pair $(w, s_i)$ is 
$$\mathcal{L}(w, s_i) = -\phi(w, s_i)+\mathrm{log}\sum_{j=0}^{|S_w|}\mathrm{exp}(\phi(w, s_j))$$

\begin{table*}[t]
    \begin{tabularx}{\textwidth}{l | c || c c c c || c c c c | c}
    \toprule
    \multirow{2}{*}{} & \textbf{Dev} & \multicolumn{4}{c ||}{\textbf{Test Datasets}} & \multicolumn{5}{c} {\textbf{Concatenation of all Datasets}} \\
    & SE07 & SE2 & SE3 & SE13 & SE15 & Nouns & Verbs & Adj. & Adv. & ALL \\
    \hline
    \hline
    \multicolumn{11}{l}{\textbf{Baseline Systems}} \\
    \hline
    WordNet S1 & 55.2 & 66.8 & 66.2 & 63.0 & 67.8 & 67.6 & 50.3 & 74.3 & 80.9 & 65.2 \\
    MFS (in training data) & 54.5 & 65.6 & 66.0 & 63.8 & 67.1 & 67.7 & 49.8 & 73.1 & 80.5 & 65.5 \\
    BERT-base & 68.6 & 75.9 & 74.4 & 70.6 & 75.2 & 75.7 & 63.7 &  78.0 & 85.8 & 73.7\\
    \hline
    \hline
    \multicolumn{11}{l}{\textbf{Prior Work}} \\
    \hline
    HCAN & - & 72.8 & 70.3 & 68.5 & 72.8 & 72.7 & 58.2 & 77.4 & 84.1 & 71.1 \\
    EWISE & 67.3 & 73.8 & 71.1 & 69.4 & 74.5 & 74.0 & 60.2 & 78.0 & 82.1 & 71.8 \\
    GLU & 68.1 & 75.5 & 73.6 & 71.1 & 76.2 & - & - & - & - & 74.1 \\
    LMMS & 68.1 & 76.3 & 75.6 & 75.1 & 77.0 & - & - & -& - & 75.4 \\
    SVC & - & - & - & - & - & - & - & - & - & 75.6 \\
    GlossBERT & 72.5 & 77.7 & 75.2 & 76.1 & 80.4 & 79.8 & 67.1 & 79.6 & 87.4 & 77.0 \\
    \hline
    \hline
    BEM & \textbf{74.5} & \textbf{79.4} & \textbf{77.4} & \textbf{79.7} & \textbf{81.7} & \textbf{81.4} & \textbf{68.5} & \textbf{83.0} & \textbf{87.9} & \textbf{79.0} \\
    \toprule
    \end{tabularx}
    \caption{F1-score (\%) on the English all-words WSD task. \textbf{ALL} is the concatenation of all datasets, including the development set \textbf{SE07}. We compare our bi-encoder model (BEM) against the WordNet S1 and most frequent sense (MFS) baselines, as well as a frozen BERT-base classifier and recent prior work on this task.}
    \label{main-results-table}
\end{table*}

\section{Experimental Setup}
\subsection{WSD Task and Datasets}
We evaluate our BEM system with the WSD framework established in \citet{raganato2017word}. We train our model on SemCor, a large dataset manually annotated with senses from WordNet that contains 226,036 annotated examples covering 33,362 separate senses \cite{miller1993semantic}. We use the SemEval-2007 (\textbf{SE07}) dataset as our development set \cite{pradhan2007semeval}; we hold out Senseval-2 (\textbf{SE2}; \citet{palmer2001english}), Senseval-3 (\textbf{SE3}; \citet{snyder2004english}), SemEval-2013 (\textbf{SE13}; \citet{navigli2013semeval}), and SemEval-2015 (\textbf{SE15}; \citet{moro2015semeval}) as evaluation sets, following standard practice. All sense glosses used in our system are retrieved from WordNet 3.0~\cite{miller1995wordnet}. 

\subsection{Baselines}
\label{baselines-section}
We compare the BEM against a number of baseline systems. We first consider two knowledge-based baselines: \textbf{WordNet S1}, which labels each example with its first (most common) sense as specified in WordNet, and most frequent sense (\textbf{MFS}), which assigns each word the most frequent sense it occurs with in the training data. 

We also compare against the pretrained model used to initialize our BEM system, \textbf{BERT-base} \cite{devlin2019bert}, by learning a linear classifier for WSD on top of frozen BERT representations output by the final layer. We learn the weights of this output layer by performing a softmax over the possible candidate senses of the target word and masking out any unrelated senses. We find that fine-tuning BERT-base on WSD classification does not improve performance over the frozen model; this finding holds for each of the pretrained encoders we consider. Specific training details for the frozen BERT baseline are given in Section \ref{model-details-section}. Since this baseline uses a standard, discrete classification setup, it backs off to the WordNet S1 predictions for unseen words.

Finally, we compare performance to six recent state-of-the-art systems. 
The \textbf{HCAN} \cite{luo2018leveraging} model incorporates sense glosses as additional inputs into a neural WSD classifier. The \textbf{EWISE} model pretrains a gloss encoder against graph embeddings before freezing the learned sense embeddings and training an LSTM encoder on the WSD task \cite{kumar2019zero}. \citet{hadiwinoto2019improved} investigates different ways of using the (frozen) pretrained BERT model to perform WSD, with their \textbf{GLU} model performing best; \citet{vial2019sense} used various sense vocabulary compression (\textbf{SVC}) approaches to improve WSD learning.\footnote{For this work, we report the best result from a comparable setting (i.e., from a single model on the same training data).} The \textbf{LMMS} system performs k-NN on word representations produced BERT against a learned inventory of embeddings for WordNet senses \cite{loureiro2019language}. \textbf{GlossBERT} fine-tunes BERT on WSD by jointly encoding the context sentences and glosses  \cite{huang2019glossbert}; this approach relies on a single, cross-encoder model, rather than our more efficient bi-encoder approach to independently encode contexts and glosses.

\subsection{Model Architecture and Optimization}
\label{model-details-section}
Our pretrained baseline is learned using a single linear layer and softmax on the output of the final layer of the frozen BERT-base model. Similarly, each encoder in the bi-encoder model is initialized with BERT-base. We obtain representations from each encoder by taking the outputs from the final layer of each encoder, and we optimize the model with a cross-entropy loss on the dot product score of these representations.\footnote{We initialize the models with BERT-base due to better baseline performance on WSD than RoBERTa-base, see Section \ref{model-ablation-section} for more details} Additional hyperparameter and optimization details are given in the supplementary materials.

\section{Evaluation}
We present a series of experiments to evaluate our bi-encoder WSD model. We first compare the BEM against several baselines and prior work on English all-words WSD (Section \ref{overall-results-section}), and then evaluate performance on the most frequent (MFS), less frequent (LFS), and zero-shot examples (Section \ref{lfs-results-section}). 

\subsection{Overall Results}
\label{overall-results-section}
Table \ref{main-results-table} shows overall F1 results on the English all-words WSD task~\cite{raganato2017word}. Frozen BERT-base is a strong baseline, outperforming all of the prior work that does not incorporate pretraining into their systems (GAS$_{ext}$, HCAN, and EWISE). The GLU and SVC systems, which use the representations learned by BERT without fine-tuning, both slightly outperform our pretrained baseline. GlossBERT achieves even better WSD performance by fine-tuning BERT with their cross-encoder approach.

However, we also find that our BEM achieves the best F1 score on the aggregated ALL evaluation set, outperforming all baselines and prior work by at least 2 F1 points. This improvement holds across all of the evaluation sets in the WSD evaluation framework as well as for each part-of-speech on which we perform WSD. Therefore, we see that although many of the prior approaches considered build on pretrained models, we empirically observe that our bi-encoder model is a particularly strong method for leveraging BERT.

\begin{table}[t]
\centering
    \resizebox{\columnwidth}{!}{
    \begin{tabular}{l| c c | c c }
    \toprule
    & \multirow{2}{*}{\textbf{MFS}} & \multirow{2}{*}{\textbf{LFS}} & \multicolumn{2}{c}{\textbf{Zero-shot}}\\
    & & & \textbf{Words} & \textbf{Senses} \\
    \hline
    WordNet S1 & 100.0 & 0.0 & 84.9 & 53.9 \\
    BERT-base & 94.9 & 37.0 & 84.9 & 53.6 \\
    \hline
    EWISE & 93.5 & 31.2 & 91.0 & - \\
    \hline
    BEM & 94.1 & 52.6 & 91.2 & 68.9 \\
    BEM-bal & 89.5 & 57.0 & 91.9 & 71.8 \\
    \toprule
    \end{tabular}}
    \caption{F1-score (\%) on the MFS, LFS, and zero-shot subsets of the \textbf{ALL} evaluation set. Zero-shot examples are the words and senses (respectively) that do not occur in the training data. The balanced BEM system, BEM-bal, is considered in Section \ref{balanced-model-section}.}
    \label{lfs-results-table}
\end{table}

\begin{table}[t]
    \centering
    \begin{tabular}{l | c c}
    \toprule
    \textbf{Model Ablation} & \textbf{Dev F1} & $\Delta$ \\
    \hline
    Full BEM & 74.5 & - \\
    \hline 
    Frozen Context Encoder & 70.1 & -4.4\\
    Frozen Gloss Encoder & 68.1 & -6.4 \\
    Tied Encoders & 74.1 & -0.4 \\
    \toprule
    \end{tabular}
    \caption{Ablations on the bi-encoder model (BEM). We consider the effect of freezing each of the two encoders and of tying the parameters of the encoders on development set performance.}
    \label{ablations-table}
\end{table}

\subsection{Zero-shot and Rare Senses Results}
\label{lfs-results-section}
To better understand these overall results, we break down performance across different sense frequencies. We split examples from the aggregated ALL evaluation set into mentions with the most frequent sense (MFS) of the target word and mentions that are labeled with the other, less frequent senses (LFS) of that word. 
We also consider zero-shot performance for both unseen words and unseen senses by evaluating performance on examples that are not observed during training. We compare our model against the frozen BERT-base baseline and EWISE \cite{kumar2019zero}, which also reported performance in these settings (Table \ref{lfs-results-table}).

\paragraph{BEM performs best on rare senses.} The vast majority of BEM's gains comes from better performance on the LFS examples, leading to a 15.6 F1 improvement over the BERT baseline on the LFS subset. Despite this gain on less frequent senses, BEM remains (approximately) as accurate on the MFS examples as prior work and the BERT baseline. While we still see a large difference of 41.5 F1 points between the MFS and LFS examples with BEM, this is a strong improvement over both the BERT-base baseline and the EWISE system. 

\paragraph{BEM shows competitive performance on unseen words.} Next we evaluated BEM on zero-shot words that did not occur in the training data. In this setting, WordNet S1 is a very strong baseline that achieves almost 85 F1 points from an untrained knowledge-based approach. Since the BERT-base model backs off to the WordNet S1 baseline for unseen words, it gets the same performance in this setting. The EWISE model from previous work, as well as our BEM, both outperform this baseline, with the BEM achieving a slightly higher F1 score for zero-shot words. 

\paragraph{BEM generalizes well to embedding zero-shot senses.}
The bi-encoder model allows us to predict senses that do not occur in the training set by embedding senses; 
this is a valuable modeling contribution since many senses do not occur in even the largest manually labeled WSD datasets. We therefore evaluate the BEM and baselines on zero-shot senses. The WordNet most common sense baseline remains strong, and the BERT baseline performs similarly to this WordNet S1 baseline. However, our bi-encoder model outperforms both baselines by at least 15 F1 points. This demonstrates that BEM is able to learn useful sense representations from the gloss text that are able to generalize well to unseen senses.

\section{Analysis Experiments}
In our model evaluation, we found that BEM outperforms prior work by improving disambiguation of less frequent senses while maintaining high performance on common ones. This section presents a series of analysis experiments in order to determine which aspects of the approach contribute to these improvements. In Section \ref{model-ablation-section}, we ablate different aspects of our model, and we consider the effect of balancing the training signal across senses with different frequencies in Section \ref{balanced-model-section}. Finally, we perform a qualitative analysis of the learned sense embedding space in Section \ref{sense-embedding-section}. 

\begin{table}[t]
    \centering
    \begin{tabular}{l | c}
    \toprule
    \textbf{Pretrained Model} & \textbf{Dev F1} \\
    \hline
    BERT-base & 68.6 \\
    BERT-large & 67.5 \\
    RoBERTa-base & 68.1 \\
    RoBERTa-large & 69.5 \\
    \toprule
    \end{tabular}
    \caption{Performance of various pretrained encoders on the WSD development set.}
    \label{pretrained-probe-table}
\end{table}

\begin{figure*}[t]
\centering
\includegraphics[scale=0.25]{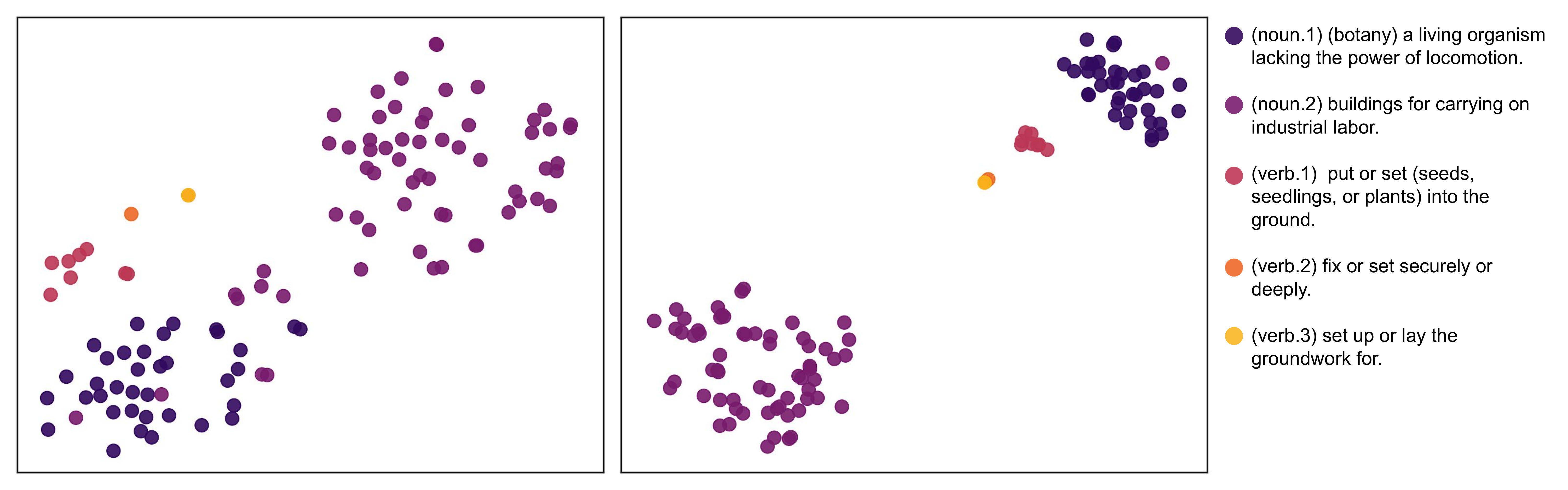}
\caption{Representations for the word \textit{plant} encoded by the frozen BERT-base encoder (left) and the context encoder of our BEM system (right); visualized with t-SNE. Sense glosses are from Wordnet \cite{miller1995wordnet}.}
\label{sense-scatter-plot}
\end{figure*}

\subsection{Model Ablations}
\label{model-ablation-section}
We ablate aspects of the bi-encoder model in order to see how they contribute to the overall performance; we consider freezing the context encoder, freezing the gloss encoder, and tying the two encoders so that they share the same parameters.

The results are shown in Table \ref{ablations-table}. A frozen gloss encoder hinders the system more than a frozen context encoder, implying that the gloss encoder needs to update the pretrained parameters more than the context encoder. We also see that while having independent encoders gives us the best performance, tying the parameters of the two encoder harms performance much less than freezing either of them. The tied encoder ablation leads to a 0.4 F1 point decrease on SemEval2007, and outperforms all prior models on this evaluation set despite having half the trainable parameters of the full BEM system.  

Next, we consider how the choice of pretrained model affects WSD performance. Table \ref{pretrained-probe-table} shows the performance of \textbf{BERT-base} and \textbf{BERT-large} \cite{devlin2019bert} on the WSD SemEval2007 evaluation set, which is used as our development set; we also consider the WSD performance of \textbf{RoBERTa-base} and \textbf{RoBERTa-large} \cite{liu2019roberta}. Similarly to the pretrained BERT-base baseline from previous section, we do not fine-tune the pretrained encoders, as we found that for all considered pretrained encoders that this did not improve performance over the frozen model.

Surprisingly, we see similar performance on the development set across all of the encoders we consider, despite the large pretrained models having twice as many parameters as the base models. Although RoBERTa-large does slightly outperform the BERT-base encoder, we initialize the BEM with BERT-base for better training efficiency.

\subsection{Balancing the Senses}
\label{balanced-model-section}
Despite the improvement on less common senses over baselines (Section \ref{lfs-results-section}), we still see a large performance gap between the MFS and LFS subsets. One possible explanation is data imbalance, since the MFS subset contains many more training examples. To control for this effect, we consider an additional training scheme for the bi-encoder model, in which we re-balance the training signal for each candidate sense of a target word. We do this by weighting the loss of each sense $s$ in the set of candidate senses of the target word $w$ by its inverse frequency in the training data. By doing this, we allow each sense to contribute equally to the training signal for $w$.

This balanced BEM model achieves an F1 score of 77.6, underperforming the standard BEM on the aggregated ALL evaluation set. Table \ref{lfs-results-table} shows the performance of the balanced BEM. By breaking down the balanced model performance, the balanced BEM outperforms the standard BEM on LFS examples, but suffers from worse performance on the more common MFS examples. We also find that this balancing during training slightly improves performance on both zero-shot words and senses.

These findings show that while weighting the data gives better signal for less common senses, it comes at the cost of the (sometimes helpful) data bias towards more frequent sense. This finding holds with similar results from \citet{postma2016more}, although their experiments focused on altering the composition of the training data, rather than modifying the loss. One possible direction for future work is a more thorough investigation of methods for obtain a stronger training signal from less frequent senses, while still taking the MFS bias into account.

\subsection{Visualizing Sense Embeddings}
\label{sense-embedding-section}
Finally, we explore the word representations learned by our bi-encoder model from fine-tuning on the WSD task. We perform a qualitative evaluation on the representations output by the BEM context encoder and compare these representations against those from the final layer of the frozen BERT-base encoder. 

Figure \ref{sense-scatter-plot} shows the outputs from each system on all instances of the word \textit{plant} in the SemCor dataset. We see that BERT-base already learns some general groupings of the senses without any explicit word sense supervision; however, the sense clusters become much more concentrated in the bi-encoder model. We also see that the noun senses are better separated by the BEM than the verb senses (which all cluster near each other). This is most likely due to the limited training data for these verb senses compared to the much more common noun sense examples. We present additional visualizations of other ambiguous words in Appendix~\ref{supplement-exploration}.

\section{Few-shot Learning of WSD}
In this section, we investigate how efficient the BEM is in a few-shot learning setting, by limiting the number of training examples the model can observe per sense. We hypothesize that our model will be more efficient than a standard classifier for learning WSD, due to the additional information provided by the sense definitions.

In order to simulate a low-shot data setting, we create $k$-shot training sets by filtering the SemCor data such that the filtered set contains up to $k$ examples of each sense in the full dataset; we then train the bi-encoder model using only this filtered training data. We train models on values of $k=1, 3, 5, 10$ and compare their performance against the model trained on the full train set. We also retrain the frozen BERT-base classifier baseline for each $k$ considered. In order to keep training comparable across different amounts of training data, we train each few-shot BEM for the same number of training steps as the system trained on the full dataset (approximately 180,000 updates).

\begin{figure}[t]
\centering
\includegraphics[width=0.9\linewidth]{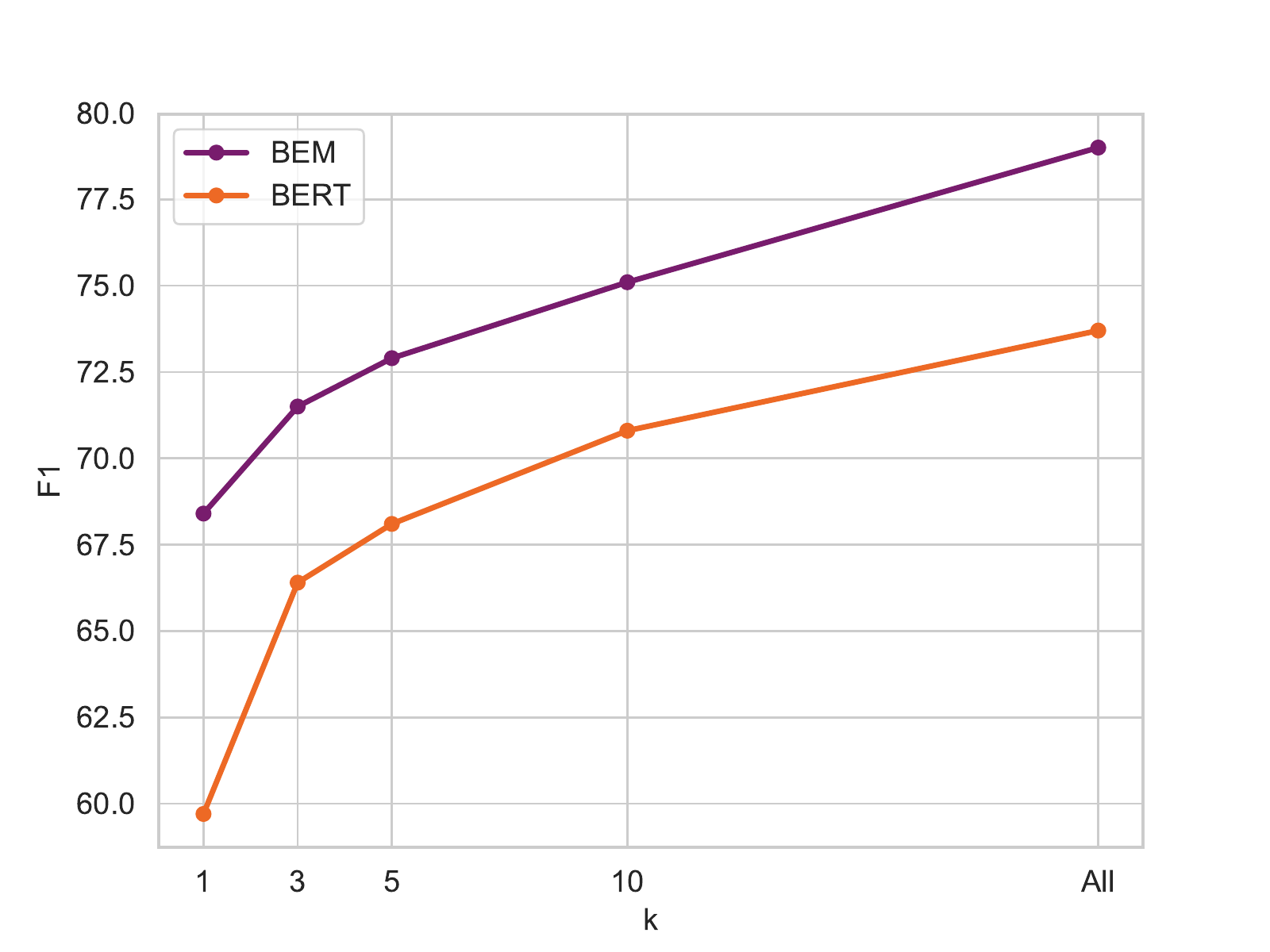}
\caption{Performance of WSD models on the \textbf{ALL} evaluation set, trained in the few-shot setting across different values of $k$ and compared against the systems trained on the full training set ($k$ = All).}
\label{fewshot-plot}
\end{figure}

The results of this experiment are given in Figure \ref{fewshot-plot}. Unsurprisingly, both the frozen BERT classifier and the BEM achieve better F1 scores as we increase $k$ and train them on additional data. However, we see that the BEM is more efficient on smaller values of $k$, with a much smaller drop off in performance at k=1 than the pretrained baseline. This efficiency also allows the BEM to achieve similar performance to the full baseline model with only 5 (or fewer) examples per sense. 

The performance of these few-shot models gives us insight into the the kinds of data that could be used to improve WSD models. While it would be prohibitively difficult to annotate many examples for every sense considered by a WSD system, it is possible that augmenting existing WSD data to provide a few labeled examples of rare senses could be more effective than simply annotating more data without considering the sense distribution. These sorts of considerations are particularly important when extending the WSD task to new domains or languages, where a great deal of new data needs to be annotated; an important goal for these sorts of data augmentation is to make sure they allow for the efficient learning of \textit{all} senses.

\section{Conclusion}
In this work, we address the issue of WSD systems underperforming on uncommon senses of words. We present a bi-encoder model (BEM) that maps senses and ambiguous words into the same embedding space by jointly optimizing the context and glosses encoders. The BEM then disambiguates the sense of each word by assigning it the label of the nearest sense embedding. This approach leads to a 31.1\% error reduction over prior work on the less frequent sense examples.

However, we still see a large gap in performance between MFS and LFS examples, with our model still performing over 40 points better on the MFS subset. Most recent WSD systems show a similar trend: even the representations of frozen BERT-base that are not fine-tuned on WSD can achieve over 94 F1 on examples labeled with the most frequent sense. 

This leaves better disambiguation of less common senses as the main avenue for future work on WSD. Potential directions include finding ways to obtain more informative training signal from uncommon senses, such as with different approaches to loss reweighting, and exploring the effectiveness of other model architectures on LFS examples. Another direction for future work would improve few-shot approaches to WSD, which is both important for moving WSD into new domains and for modeling rare senses that naturally have less support in WSD data.

\section*{Acknowledgements}
This material is based on work conducted at the University of Washington, which was supported by the National Science Foundation Graduate Research Fellowship Program under Grant No. DGE-1762114.
We thank Gabi Stanovsky, Ledell Wu, and the UW NLP group for helpful conversations and comments on the work.

\bibliography{acl2020}
\bibliographystyle{acl_natbib}

\appendix
\section{Additional Training Details}
Both our frozen BERT baseline and the BEM are implemented in PyTorch\footnote{https://pytorch.org/} and optimized with Adam \cite{kingma2015adam}. The pretrained models used to initialize each model are obtained through \citet{Wolf2019HuggingFacesTS}; we initialize every model with the \textit{bert-base-uncased} encoder. 
 
\paragraph{BERT-base baseline.} The linear layer of the frozen BERT-base classifier is trained for 100 epochs, and tuned over the following parameter ranges: learning rates of $[5e-6, 1e-5, 5e-5, 1e-4]$ and batch sizes of $[32, 64, 128]$. 

\paragraph{Bi-encoder Model (BEM).} The BEM is trained for 20 epochs with a warmup phase of 10,000 steps. We use a context batch size of 4 and a gloss batch size of 256. The model is tuned on learning rates in $[1e-6, 5e-6, 1e-5, 5e-5]$. We use two GPUs to train the BEM, optimizing each encoder on a separate GPU to allow for larger batch sizes.

\section{Additional Sense Embedding Explorations}
\label{supplement-exploration}
\begin{figure*}[t]
\centering
\includegraphics[scale=0.25]{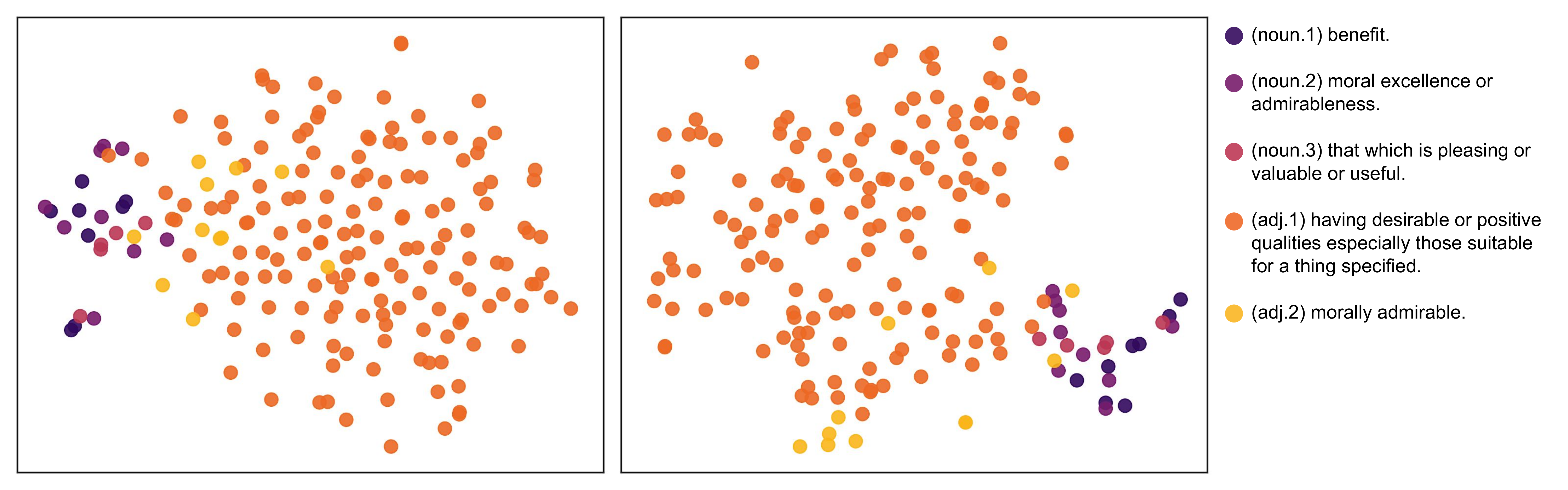}
\caption{Visualization of learned representations for the word \textit{good}. Overall the BEM (right) doesn't improve on the frozen BERT representations (left), but we observe that the \textit{adj.2} sense is becoming better distinguished from \textit{adj.1} by the BEM, with the examples of \textit{adj.2} appearing only in one edge of the cluster for \textit{adj.1}.}
\label{sup-good-scatter}
\end{figure*}

\begin{figure*}[t]
\centering
\includegraphics[scale=0.25]{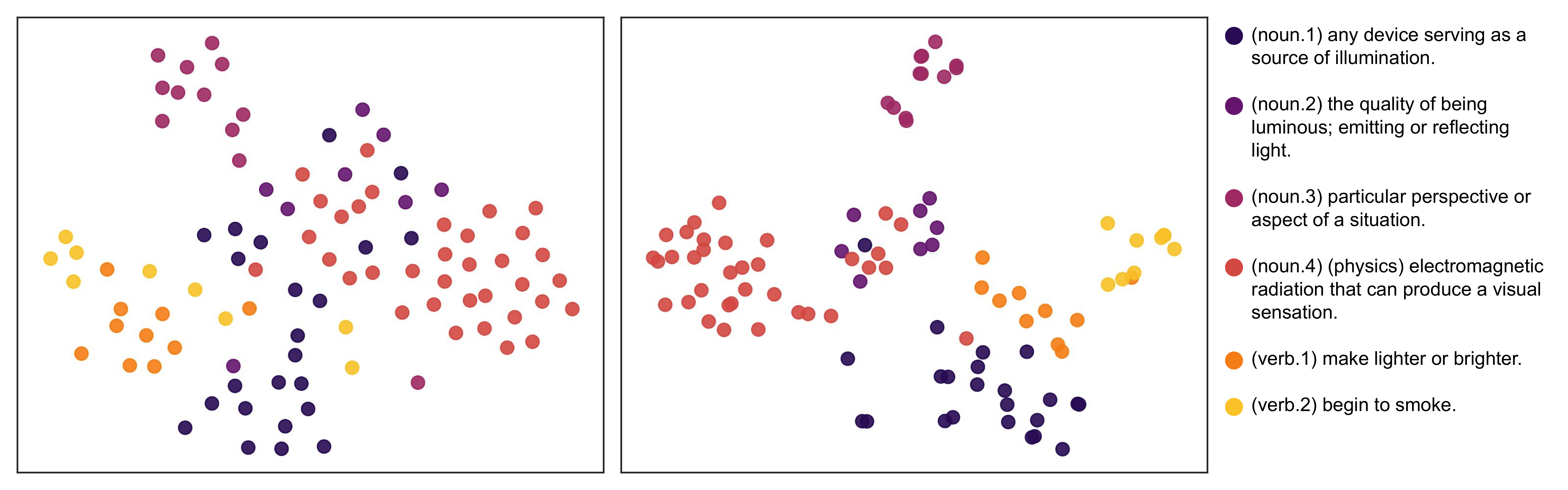}
\caption{Visualization of learned representations for the word \textit{light}. We see more distinct clusters forming in the representations from the BEM (right) than in the BERT-base outputs (left), though there is still overlap with the edges of the some groups.}
\label{sup-light-scatter}
\end{figure*}

\begin{figure*}[t]
\centering
\includegraphics[scale=0.25]{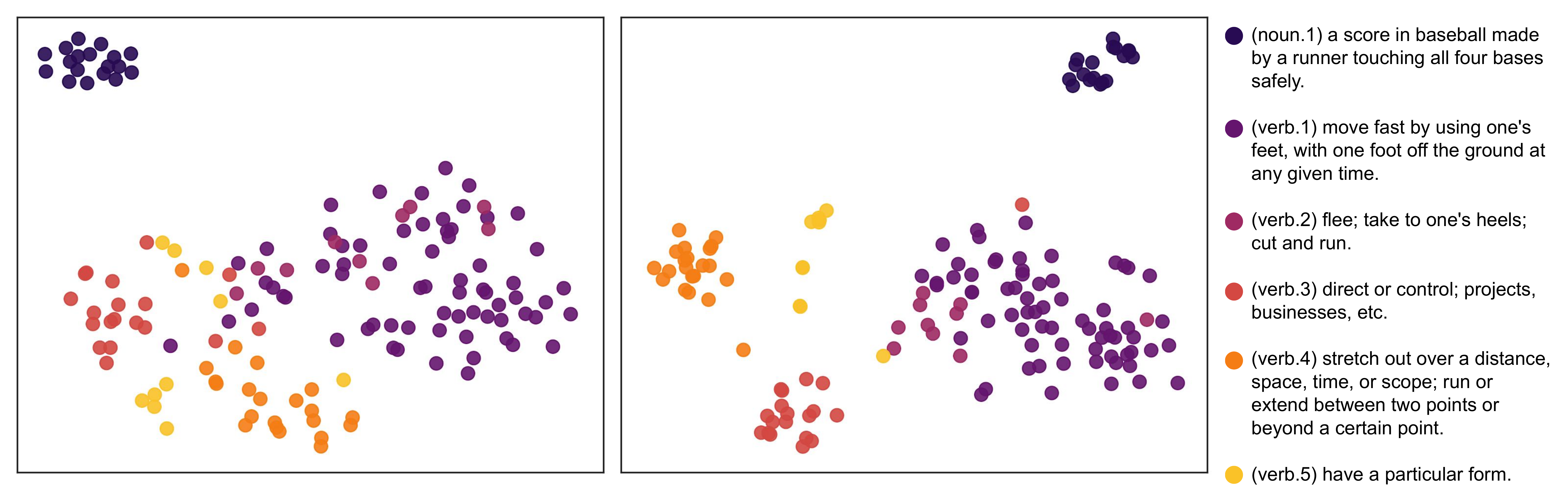}
\caption{Visualization of learned representations for the word \textit{run}.  We see that both the frozen BERT model (left) and BEM system (right) has difficulty distinguishing the \textit{verb.1} and \textit{verb.2} senses of \textit{run}, which are closely related senses with a very fine-grained distinction (see glosses given in legend).}
\label{sup-run-scatter}
\end{figure*}

We present additional sense embedding space visualizations (Figures \ref{sup-good-scatter}, \ref{sup-light-scatter}, and \ref{sup-run-scatter}). These visualizations are generated identically to the one discussed in Section \ref{sense-embedding-section}. In each figure, the left visualization shows the representations output by a frozen BERT-base model, and the right one shows the output of our BEM's context encoder. All figures are visualized with t-SNE. We choose words from SemCor that occur more than 50 times; for clarity, we limit the visualization to the six most common senses of each word. All senses and glosses are gathered from WordNet \cite{miller1995wordnet}.

\end{document}